\documentclass[runningheads]{llncs}

\usepackage{eccv}

\usepackage{eccvabbrv}

\usepackage{graphicx}
\usepackage{booktabs}

\usepackage[accsupp]{axessibility}  
\newcommand{\ttb}[1]{{\fontfamily{lmtt}\fontseries{b}\selectfont #1}}

\usepackage{hyperref}

\usepackage{orcidlink}
\usepackage{wrapfig}
\usepackage{enumitem}
\usepackage{multirow}
\usepackage{pifont}

\begin{document}

\title{Attribute Token Arithmetic: Disentangled and Continuous Semantic Control for Visual Autoregressive Models} 

\titlerunning{ATA}

\author{Xindi Yang \orcidlink{0009-0003-7287-9843} \and
Yicheng Wu \orcidlink{0000-0002-7669-9167}\thanks{Corresponding author} \and
Cheng Zhang \orcidlink{0000-0003-3145-9036} \and
\\ Jianfei Cai \orcidlink{0000-0002-9444-3763} \and
Tien-Tsin Wong \orcidlink{0000-0002-7792-9307}}

\authorrunning{X.~Yang et al.}

\institute{Faculty of IT, Monash University \\
\email{
\{xindi.yang, yicheng.wu1, cheng.zhang, jianfei.cai, tt.wong\}@monash.edu}
}

\maketitle

\begin{abstract}
  Autoregressive text-to-image generation has recently achieved remarkable progress, offering high-fidelity synthesis via a unified generative framework. However, fine-grained semantic control remains challenging due to the attribute entanglement and the misalignment between textual and fine-grained visual representations. In this paper, we introduce Attribute Token Arithmetic (ATA), a method that enables disentangled and continuous attribute control in visual autoregressive modelling. Inspired by the vector arithmetic property observed in word embeddings, ATA identifies semantic directions corresponding to visual attributes (\eg, aging, fatness, emotion) directly within the pretrained autoregressive latent space. These directions are learned from a single reference image, without model retraining or large-scale supervision. During generation, attributes can be continuously adjusted and compositionally combined through simple arithmetic operations with other attribute tokens. Extensive experiments demonstrate that ATA achieves identity-preserving, fine-grained, and multi-attribute adjustment, outperforming existing autoregressive editing baselines in controllability, generality, and computational efficiency. Our code will be released \href{https://github.com/Madaoer/ATA}{here}.
\end{abstract}

\section{Introduction}
\label{sec:intro}

Photorealistic image synthesis and editing are fundamental tasks in visual content creation, underpinning applications in movie production, gaming, and virtual reality.
Traditional computer graphics achieve these goals through geometric modeling and physics-based rendering, but such pipelines are labor-intensive and require substantial expertise.
Recent years have witnessed a paradigm shift toward generative AI, which produces realistic images directly from data, significantly lowering the barrier to visual content creation~\cite{rombach2022high,xing2024dynamicrafter,yang2024cogvideox,Yang_2025_ICCV,wan2025wan,wu2026codebrain,zheng2024open}.

Among generative paradigms, diffusion models~\cite{rombach2022high,song2020score,song2020denoising,Wu_2024_CVPR} have achieved remarkable success in text-to-image generation, enabling high-fidelity and semantically aligned visual synthesis. For example, TokenVerse~\cite{garibi2025tokenverse} and XVerse~\cite{chen2025xverse} attempt to achieve concept-level control by associating text tokens with visual semantics, allowing multi-concept manipulation.
However, diffusion models require iterative denoising steps, which are computationally expensive and limit their efficiency for real-time or high-resolution editing~\cite{song2020denoising,lu2022dpm}.
This limitation has renewed interest in autoregressive (AR) models, which generate images token by token in a discrete latent space, offering faster sampling, superior scalability, and a unified formulation with language generation~\cite{sun2024autoregressive,chen2025janus}.

\begin{figure}[tb]
  \centering
  \includegraphics[width=\textwidth]{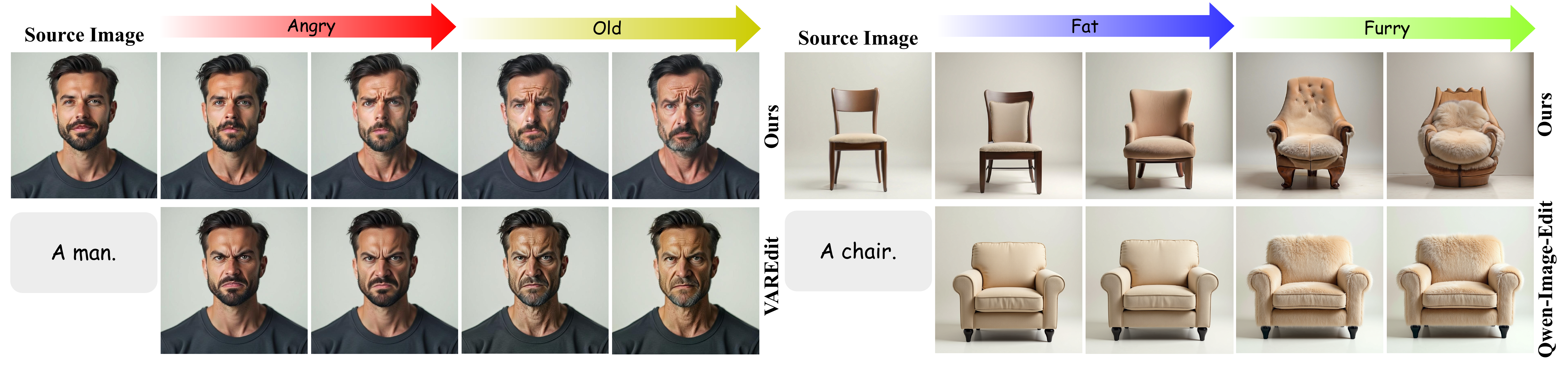}
  \caption{\textbf{Exemplar edited images with disentangled and continuous semantic control}. Given a reference and paired text prompts, our ATA extracts semantic attribute directions in the VAR-aligned text space. These directions are derived from the image concepts that correspond to the semantic differences between paired prompts. Specifically, ATA modifies source images to gradually increase the strength of two attributes. The first two results increase with the first attribute, while in the last two, the second attribute is introduced.   (Left)  ATA enables disentangled and continuous semantic control of ``\ttb{angry}'' and ``\ttb{old}'', while preserving the identity.  (Right) Moreover, ATA can transfer the general  {\em disentangled} attribute concepts across object categories. Although ``\ttb{fat}''-ness is not a native attribute of chairs, ATA can still transfer the ``\ttb{fat}'' \& ``\ttb{furry}'' concepts and enable continuous control to yield creative chair designs, while both VAREdit and Qwen-Image-Edit fail to provide continuous control.
  }
  \label{fig:teaser}
\end{figure}

Building upon this idea, Visual Autoregressive Models (VAR)~\cite{tian2024visual,han2025infinity} extend AR generation to high-resolution imagery through multi-scale tokenization and causal transformers.
By formulating image synthesis as a sequence prediction task, VAR naturally bridges language and vision within a discrete latent space.
Despite their success in image quality and efficiency, precise and flexible editing remains a major challenge, as fine-grained semantic control remains challenging to realize within discrete token spaces.
Such fine-grained control is essential for many graphics applications to allow users to fine-tune the desired visual effect.
Recent works attempt to address this limitation.
AREdit~\cite{wang2025training} manipulates internal representations of pretrained VARs without retraining but supports only single-attribute edits.
VAREdit~\cite{mao2025visual} fine-tunes VARs on paired examples for instruction-based editing but suffers from limited generalization.
Overall, existing approaches struggle to achieve generic, continuous, and compositional control within the VAR framework.

Therefore, our work starts from a simple yet powerful question: \textit{Can attribute-level control be achieved within VARs similar to text token arithmetic?} 
We discover that semantically meaningful vector operations exist in the VAR-aligned text–token space, where manipulating token vectors associated with specific attributes (\eg, ``\ttb{angry}'', “\ttb{old}”) produces coherent and interpretable visual changes (Fig.~\ref{fig:teaser}).
This insight reveals that text tokens inherently encode compositional visual semantics, suggesting a new paradigm for continuous and cross-category control, without any retraining.

\begin{figure}[tb]
  \centering
  \includegraphics[width=\textwidth]{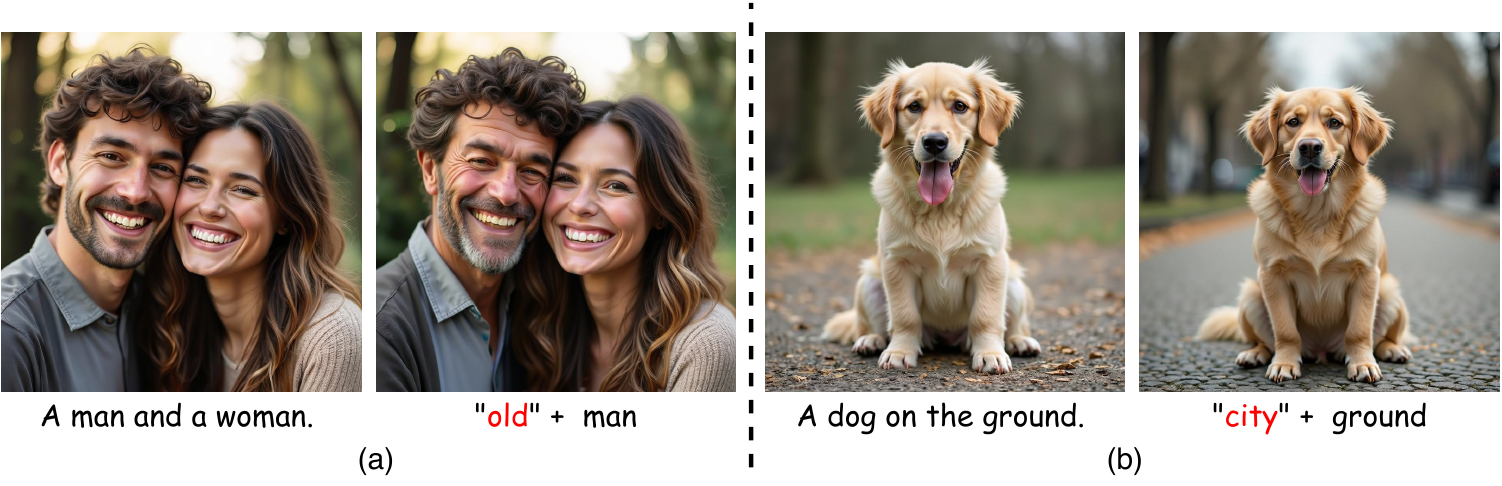}
  \caption{\textbf{Preliminary image editing results}. By manipulating semantic directions to the corresponding text tokens, (a) we achieve the localized editing and the preservation of object identities; (b) we can also edit non-localized attributes, \eg, background.
  }
  \label{fig:text_edit}
\end{figure}

Motivated by this observation, we propose ATA (Attribute Token Arithmetic), a lightweight optimization-based framework that learns multi-attribute vectors directly within a pretrained VAR.
Given a single reference image and its textual description, ATA disentangles attribute-specific directions in the joint text–token space and leverages them for controllable image synthesis.
Once learned, these attribute vectors can be stored in an offline library and reused or combined at arbitrary strengths to create new attributes and visual compositions (see Fig.~\ref{fig:teaser}).
Practically, ATA enables intuitive edits such as ``\ttb{making a face angrier}'', ``\ttb{adding a smile and aging simultaneously}'', or even ``\ttb{trans\-fer\-ring the fat attribute of a cat to a chair}''.

\noindent Our contributions are as follows:
\begin{itemize}
\item We reveal that semantically meaningful vector arithmetic exists in the latent space of pretrained VARs, enabling attribute-level visual control.

\item We propose ATA, a simple yet effective framework for learning generic and compositional attribute vectors without retraining or large-scale supervision.

\item We validate ATA through extensive experiments and user studies, demonstrating its controllability and generalizability superior to existing VAR editing approaches.

\end{itemize}

\section{Related Work}

\subsection{Autoregressive Image Generation}
The success of LLMs, such as the GPT series~\cite{achiam2023gpt}, has demonstrated the scalability and versatility of the autoregressive (AR) paradigm. To adapt this to visual data, seminal works like VQ-VAE~\cite{van2017neural} and VQGAN~\cite{esser2021taming} introduce learnable codebooks to represent image patches as discrete tokens. Subsequent variants have improved codebook utilization and reconstruction via multi-codebook designs~\cite{ma2025unitok}, residual quantization~\cite{lee2022autoregressive}, and regularization~\cite{yuvector}. Additionally, leveraging semantic priors from models like CLIP~\cite{zhu2024scaling} has enhanced the alignment between quantized tokens and textual concepts.

More recently, research has shifted toward scalar-based quantization, such as Binary Scalar Quantization (BSQ)~\cite{zhaoimage}, which offers a lightweight alternative to traditional vector quantization. To mitigate information loss, studies have explored continuous or hybrid formulations: MAR~\cite{li2024autoregressive} employs a diffusion process in continuous latent space, TokenBridge~\cite{wang2025bridging} utilizes post-training quantization, and HART~\cite{tanghart} adopts a hybrid design. Despite these continuous alternatives, discrete representations remain favored for their ability to naturally unify linguistic and visual modalities.

Once tokenized, causal transformers are trained to capture sequential dependencies. While early models like DALL-E~\cite{ramesh2022hierarchical} directly modeled flattened sequences, they faced scalability and spatial consistency challenges. Later strategies introduced hierarchical and spatially aware modeling: TiTok~\cite{yu2024image} uses register tokens for semantic regions, while column-wise tokenization~\cite{dong2025equivariant} generates tokens from left to right. Visual Autoregressive Modeling (VAR)~\cite{tian2024visual} further refined this via next-scale prediction across resolutions, a paradigm recently extended by Infinity~\cite{han2025infinity} through the integration of scalar quantization.

Collectively, these advances have significantly bridged the gap between continuous visual data and discrete sequence modeling, establishing a solid foundation for autoregressive image generation with enhanced fidelity, coherence, and flexibility. However, achieving fine-grained control in AR-based generation remains an open challenge. Text conditions, while semantically expressive, often provide only coarse guidance that fails to capture intricate visual details. Consequently, the generated images may lack completeness, transferability, precision, and structural consistency across regions or attributes. These limitations underscore the need for explicit, disentangled, and semantically grounded control mechanisms that are highly demanded in the movie and gaming industries. These motivate our exploration of content control within the autoregressive framework.

\subsection{Image Editing}
Image editing is a long-standing problem in computer vision and graphics. Recent foundation models have advanced visual content generation. This progress has shifted attention to achieve fine-grained semantic control in generative model.
Diffusion-based models have explored in this area. \cite{alaluf2024cross, hertz2022prompt, patashnik2023localizing, tumanyan2023plug} manipulate the denoising process via feature injection to control target concepts in image. \cite{ruiz2023dreambooth, gandikota2024concept} propose to train LoRA~\cite{hu2022lora} modules for specific concepts to improve disentanglement and consistency. \cite{dalva2024fluxspace, garibi2025tokenverse, guerrero2024texsliders, chen2025xverse} propose edits by locating semantic directions in a pretrained latent space. \cite{wu2025qwen,mao2025ace++} leverages large-scale image–text paired data to finetune the base model and enable instruction-based editing.
These methods can be effective in fine-grained semantic editing. However, iterative sampling in the denoising process slows inference and makes real-time editing difficult.
Recently, AR image generation has gained interest due to its higher efficiency and language-like modeling paradigm. AREdit~\cite{wang2025training} manipulates representations but focuses on single-concept edits, which limits disentanglement. ~\cite{mao2025visual,mu2025editar,wu2025omnigen2} are trained on a large-scale dataset and offer instruction-based interfaces, yet they provide limited control.
Our ATA work aims to bridge this gap, providing disentangled and continuous semantic control in the VAR models.

\begin{table}[tb]
\small
\caption{\textbf{Comparison} among existing image editing methods.}
\begin{center}
\label{tab:comparison}
\begin{tabular}{l|c|c|c}
\toprule
Model & \shortstack{Strength \\ Control} & \shortstack{Cross-Category \\ Transfer} & \shortstack{Object \\ Composition} \\
\midrule
Concept Slider & \ding{51} & \ding{51} & \ding{55} \\
Tokenverse & \ding{55} & \ding{55} & \ding{51} \\
VAREdit, Qwen-Image-Edit & \ding{51} & \ding{51} & \ding{55} \\
\midrule
\textbf{Ours} & \ding{51} & \ding{51} & \ding{51} \\
\bottomrule
\end{tabular}
\end{center}
\end{table}

\textbf{Comparison with Existing Methods.} Image editing has progressed significantly in both the Diffusion and AR communities, with multiple approaches emerging to address similar objectives. Table~\ref{tab:comparison} compares our work with these recent image editing methods.

\section{VAR Basics}
VAR~\cite{tian2024visual,han2025infinity} formulates text-to-image generation as a sequence prediction task in the discrete token space. Given a text prompt, a frozen language encoder (\ie, Flan-T5~\cite{chung2024scaling}) produces text tokens. These text tokens are then mapped into a VAR-aligned text token space via an MLP network to align with image tokens.
An image tokenizer encodes the image $I$ into a feature map $F$ and then quantizes $F$ into $K$ multi-scale residual maps $(R_1, R_2, ..., R_{\rm k})$. In our experiments, there are 13 scales ranging from $1\times1$ to $64\times64$ to synthesize $1024\times1024$ images.

\begin{figure}[tb]
  \centering
  \includegraphics[width=\textwidth]{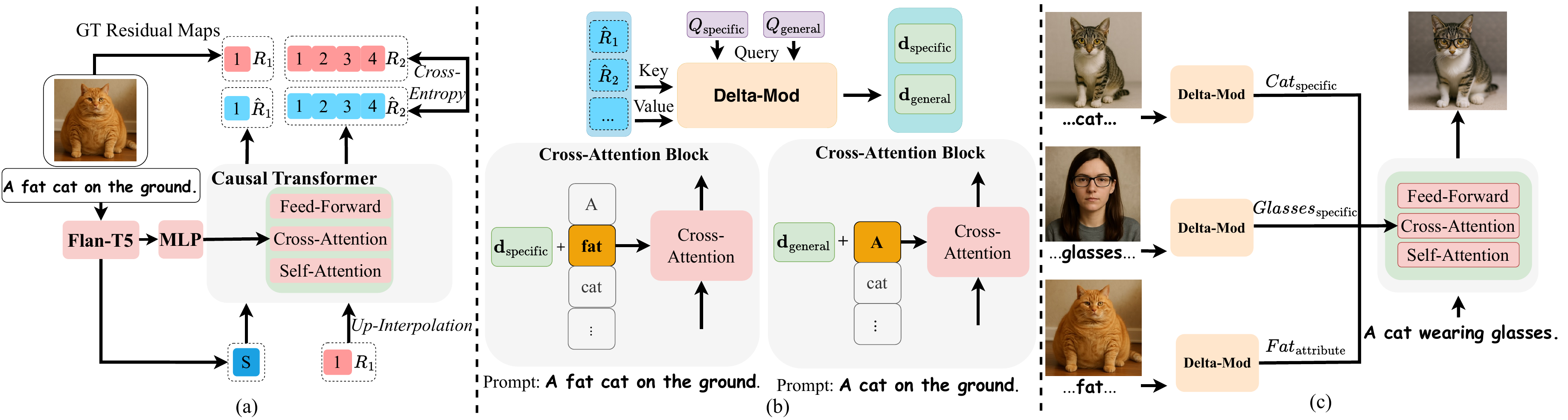}
  \caption{\textbf{Pipeline of our ATA framework}. (a) A pretrained VAR model processes image and text tokens with a causal transformer. At each next-scale prediction step, VAR applies Self-Attention, Cross-Attention, and Feed-Forward blocks. Text and image tokens interact through cross-attention. Our ATA adds the learned semantic directions to the corresponding VAR-aligned text tokens in the Cross-Attention Block. (b) Given a reference image and a pair of semantically different prompts, our ATA learns two semantic directions, $\vec{d}_{\rm general}$ and $\vec{d}_{\rm specific}$, for each target concept, via a lightweight attention block, Delta-Mod. These directions capture the general attribute concept and image-specific personalized traits. They are learned with the original VAR training loss. (c) During inference, the learned directions are applied to the corresponding text tokens. Our ATA enables disentangled and continuous control in VAR.
  }
  \label{fig:framework}
\end{figure}

We adopt Infinity~\cite{han2025infinity} as our VAR backbone. As illustrated in Fig.~\ref{fig:framework}(a), each transformer block consists of self-attention, cross-attention, and feed-forward layers, where the text tokens guide the generation of image tokens through cross-attention. During training, the model predicts next-scale residual $\hat{R}_{\rm k+1}$ conditioned on all previous residual maps $(R_1, R_2, ..., R_{\rm k})$ and VAR-aligned text context. The causal transformer takes low-resolution residual maps as input and predicts the next-resolution residual maps through a progressive up-interpolation process. 
During the inference, image generation proceeds hierarchically from coarse to fine scales. Coarse-scale tokens capture the global semantic structure, while fine-scale tokens incrementally refine spatial details until the full-resolution image is reconstructed. The fine-scale components are represented as residual maps, forming a Laplacian pyramid–like hierarchy that progressively injects local details into global semantics.

\section{Method}
\label{sec:blind}
Inspired by the vector arithmetic behavior observed in word embeddings~\cite{mikolov-etal-2013-linguistic}, we first investigate in Sec.~\ref{subsec:semdir} whether various attributes (such as fatness, aging) can be edited via controlling the semantic directions in the pretrained VAR space. We found that there indeed exist controllable semantic directions, which enable localized semantic manipulation in the VAR-aligned text token space. To achieve more fine-grained control within this space for personalized image editing, we propose to learn disentangled semantic directions from individual images in Sec.~\ref {subsec:disen_attr_learning}.

\subsection{Semantic Control in Text Tokens}
\label{subsec:semdir}

Prior work in NLP~\cite{mikolov-etal-2013-linguistic} shows that word embeddings support vector arithmetic consistent with human intuition. For example, \ttb{king + woman - man $\approx$ queen}, where simple addition and subtraction on token embeddings produce the desired output aligned with human intuition.
Inspired by the vector arithmetic behavior observed in word embeddings, we explore whether VAR supports a similar form of vector arithmetic for image editing.

In VAR’s text-to-image generation task, a frozen text encoder processes the prompt $P$ to produce text tokens, which are then mapped by a pretrained MLP to the VAR-aligned text-embedding space. These tokens interact with image tokens in the VAR transformer and guide the generation process, as shown in Fig.~\ref{fig:framework} (a).
For simplicity, consider an image of a fat cat. We can construct a prompt pair $P_{\rm src}$: ``\ttb{A cat.}'' and $P_{\rm tar}$: ``\ttb{A fat cat.}'' We can regard ``\ttb{A}'' and ``\ttb{A fat}'' as the attribute tokens $T_{\rm attr}^{\rm src}$ and $T_{\rm attr}^{\rm tar}$, respectively, while regarding ``\ttb{cat}'' as the identity token $T_{\rm id}$ in VAR-aligned text-token space. 
We treat the attribute difference between $T_{\rm attr}^{\rm src}$ and $T_{\rm attr}^{\rm tar}$ as the semantic direction:
\begin{align} \label{eq:sd}
\Delta T_{\rm attr} = {\rm mean}(T_{\rm attr}^{\rm tar}) - T_{\rm attr}^{\rm src},
\end{align}
where $\rm{mean}(\cdot)$ denotes the mean  of the multiple text tokens associated with $T_{\rm attr}^{\rm tar}$. 
With $\Delta T_{\rm attr}$, we can manipulate the semantic direction in  token space as follows:
\begin{align}
\hat{T}_{\rm attr}^{\rm src} = T_{\rm attr}^{\rm src} + \gamma \times \Delta T_{\rm attr},
\end{align}
where $\gamma$ is a hyperparameter that controls the strength of the semantic direction. By adding a specific semantic direction to the attribute token, we can edit the corresponding attribute in the image. As shown in Fig.~\ref{fig:text_edit}, the edits are highly localized and do not affect unrelated regions. The effect is not limited to attributes and objects; non-object concepts (\eg, background) can also be edited in the same way.

\subsection{Disentangled Semantic Direction Learning}
\label{subsec:disen_attr_learning}

\begin{wrapfigure}{r}{0.5\textwidth}
  \centering
  \includegraphics[width=0.48\textwidth]{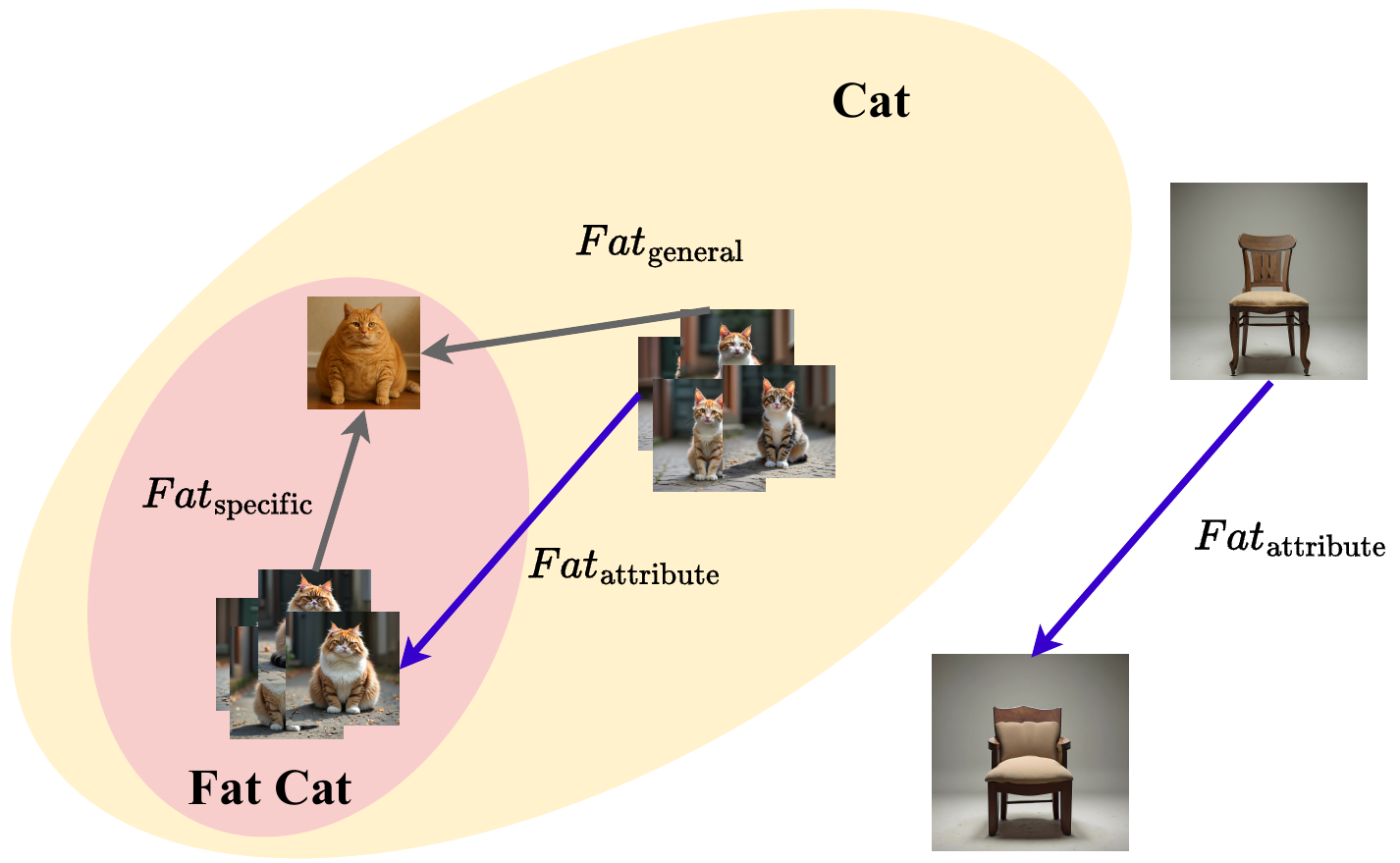}
  \caption{\textbf{Concept illustration of ATA}. Given a reference image $I$ and a pair of semantically different prompts (\eg, ``\ttb{A fat cat.}'' and ``\ttb{A cat.}''), our method learns two semantic directions with respect to $I$: $Fat_{\rm specific}$, the semantic direction from the mean of the distribution ``\ttb{fat cat}'' to $I$, and $Fat_{\rm general}$, the semantic direction from the mean of the distribution ``\ttb{cat}'' to $I$, both in the VAR-aligned text space. We obtain the general attribute direction $Fat_{\rm attribute}$ by simple vector arithmetic. The learned attribute transfers across object categories (\eg transfer "fat" to chair). }
  \label{fig:slider_compose}
\end{wrapfigure}

Using the text token embedding difference in Eq.~\eqref{eq:sd} directly as the semantic direction is insufficient to represent the general and specific concepts embodied in an image. Here, we aim to learn semantic directions from individual images, which can later be used in fine-grained personalized image editing.

\subsubsection{Disentangled Semantic Direction Representation}

Consider a text-to-image generation process, which maps the text prompt $P$ to the distribution $\mathcal{S}_P$ in the VAR-aligned text token space (corresponding to a set of images that satisfy the text prompt $P$). The model sampling process picks a sample, $\phi_I$, from the distribution $\mathcal{S}_P$, where $I$ is the specific image generated. 
To represent the semantic shift in the latent space, we establish a direction $\vec{d}$ that originates from the mean of the distribution $\mathcal{S}_P$ and points toward the target sample point $\phi_I$. This orientation is formulated as:
\begin{align}
\vec{d}_{P} &= \phi_I - \mu(\mathcal{S}_P),
\end{align}
where $\mu(\mathcal{S}_P)$ denotes the mean  of distribution $\mathcal{S}_P$ in the same latent space.

Let us use Fig.~\ref{fig:slider_compose} as an example to illustrate our key idea. Specifically, we use the attribute (\ie, \ttb{fat}) to describe a subject (\ie, \ttb{a cat}) in the text prompt, which effectively specifies a sub-distribution (\ie, \ttb{a fat cat}) out of the original distribution (\ie, \ttb{a cat}). For a specific sample $\phi_I$ within the sub-distribution, we can form two vectors: one from the mean of the original distribution towards $\phi_I$, denoted as $\vec{d}_{\rm general}$ (\ie, $Fat_{\rm general}$), and the other from the mean of the sub-distribution towards $\phi_I$, denoted as $\vec{d}_{\rm specific}$ (\ie, $Fat_{\rm specific}$). 
Note that $\vec{d}_{\rm general}$ and $\vec{d}_{\rm specific}$ can serve as semantic directions that correspond to the identity concepts in the image $I$.
By subtracting these two vectors, we effectively disentangle the {\em attribute direction} from the specific image: 
\begin{align} \label{eq:attr}
\vec{d}_{\rm attribute} &= \vec{d}_{\rm general}-\vec{d}_{\rm specific},
\end{align}
where $\vec{d}_{\rm attribute}$ (\ie, $Fat_{\rm attribute}$) identifies a generic attribute direction, from the original distribution towards the sub-distribution.
Amazingly, such general attributes can generalize across different object categories (here, transferring \ttb{fat}, learned from \ttb{fat cat},  to \ttb{chair}). 

\subsubsection{Learning Disentangled Semantic Directions} 
We now present
how to learn the disentangled semantic directions $\vec{d}_{\rm general}$ and $\vec{d}_{\rm specific}$ from a given image.
Specifically, we learn them via constructing semantically different prompt pairs.
Given a reference image $I$, we define a general prompt $P_{\rm general}$ and a specific prompt $P_{\rm specific}$ (\eg, ``\ttb{A cat on the ground.}'' and ``\ttb{A fat cat on the ground.}''), and $S(P_{\rm specific})$ is a sub-distribution of $S(P_{\rm general})$.
We learn the semantic directions via a lightweight attention block, \textbf{Delta-Mod}, as shown in Fig.~\ref{fig:framework}(b) top. 
Particularly, we extract the predicted residual map $\{\hat{R}_1, \hat{R}_2, ..., \hat{R}_k\}$ from the reference image $I$ using the visual encoder in the VAR model, which is fed into the attention block Delta-Mod as $K$ and $V$. Together with two learnable input vectors as $Q$, Delta-Mod outputs the corresponding semantic direction $\{\vec{d}_{\rm general}, \vec{d}_{\rm specific}\}$.
The process can be formulated as
\begin{align}
\{\vec{d}_{\rm general}, \vec{d}_{\rm specific}\} = {\text{Delta-Mod}}(R_1,..., R_n).
\end{align}
Note that we do not have ground truth for the semantic directions. To supervise them, as shown in Fig.~\ref{fig:framework}(b) bottom, we add $\vec{d}_{\rm general}$ and $\vec{d}_{\rm specific}$ to the corresponding text token positions of $P_{\rm general}$ and $P_{\rm specific}$, respectively, both of which are required to reconstruct the given image $I$ via the VAR model. With the VAR model frozen and the same VAR reconstruction loss, we only optimize the lightweight attention block Delta-Mod. 
Multiple semantic directions corresponding to different concepts or attributes of an image can be learned together via the same Delta-Mod. 

Figure~\ref{fig:framework}(c) illustrates our inference process. With different disentangled semantic directions extracted from different images, we can combine them for fine-grained personalized image editing by selecting either specific semantic directions $\vec{d}_{\rm specific}$ (\eg, $Cat_{\rm specific}$ and $Glasses_{\rm specific}$) or the general attribute direction Eq.~\eqref{eq:attr} (\eg, $Fat_{\rm attribute}$), and applying them on the corresponding text tokens as in Eq.~(2).
Note that due to the next-scale prediction nature in VAR, we have another hyperparameter that selects the scale at which we apply the semantic direction manipulation; see the appendix for details.
In addition, we can construct a semantic attribute library offline, from which semantic directions can be retrieved and combined on demand.

\begin{figure}[tb]
  \centering
  \includegraphics[width=\textwidth]{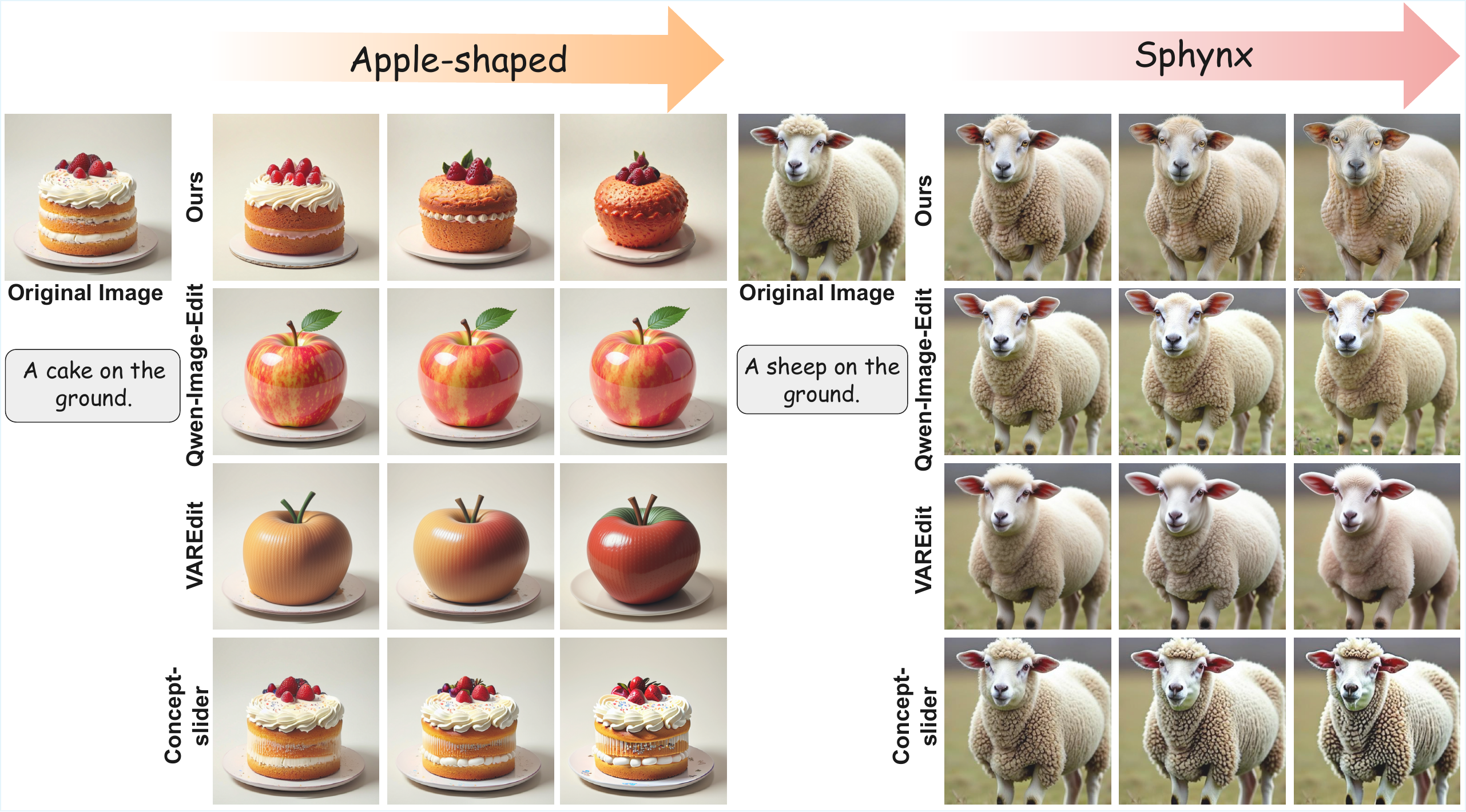}
  \caption{\textbf{General attribute transfer and cross-category semantic editing}. We present more examples of general attribute transfer. (Left) We apply the attribute ``\ttb{Apple-shaped}'' to the concept ``\ttb{cake}'' and perform continuous control. Our method gradually transforms the cake into an apple-shaped cake, whereas baseline methods turn the cake into an apple. (Right) We transfer the attribute concept ``\ttb{Sphynx}'', originally defined for Sphynx cats, to sheep. Our method achieves a gradual transition to hairlessness with pronounced skin folds, whereas the baselines fail to do so.
  }
  \label{fig:geneattr}
\end{figure}

\section{Experiments}
\label{sec:exp}

We evaluate our method against state-of-the-art baselines and demonstrate its advantages in continuous, disentangled, and generalizable semantic control. Our experiments are organized into five parts: (1) comparison with existing editing models, (2) user studies, (3) quantitative evaluation, (4) ablation studies and (5) additional qualitative demonstrations of advanced editing capabilities.

\subsection{Benchmark}
\noindent\textbf{Comparisons.}    Instruction-driven editing models follow textual commands but lack explicit mechanisms for continuous semantic control. 
We approximate continuous editing by prompting these methods with three attribute strengths: \textit{slight}, \textit{moderate}, and \textit{extreme}.
We include: \textbf{Qwen-Image-Edit (20B)~\cite{wu2025qwen}}, a large-scale DiT-based instruction editor, and \textbf{VAREdit (Infinity-8B)~\cite{han2025infinity}}, a VAR-based instruction-following editor. Furthermore, we compare with \textbf{Concept Sliders}~\cite{gandikota2024concept}, which learn continuous attribute directions by optimizing LoRA modules for each attribute concept. Here, we use their SDXL-trained~\cite{podell2023sdxl} LoRAs in our experiments.

\noindent\textbf{Data Details.}
To assess continuous multi-attribute manipulation, we construct a controlled benchmark consisting of:
\begin{itemize}[leftmargin=10pt]
    \item two object categories: humans and common objects,
    \item 15 prompts per category, each with two seeds,
    \item six attribute sets, spanning common and general concepts (\eg, \textit{apple-shaped}).
\end{itemize}
Each source image is edited with two attributes at three intensity levels.  
We then report VQA-Score~\cite{lin2024evaluating} for semantic alignment and LPIPS (VGG)~\cite{zhang2018unreasonable} for content preservation.

Moreover, we leverage the GEdit~\cite{liu2025step1x} benchmark for evaluations. We randomly select 52 different attribute-editing samples. Then, Gemini~\cite{team2023gemini} is used to generate attribute prompt pairs in the form of ``a [attribute] [object] [context]'' vs. ``a [object] [context]'', sharing the same context, along with the corresponding reference image. ATA is used to obtain general semantic attribute directions. In this way, we built an attribute library in GEdit.
Beyond semantic consistency and perceptual quality in  GEdit, we introduce a disentanglement score to monitor unintended changes in [attribute2] when [attribute1] is modified. For the evaluation of GEdit, we employ the Qwen2.5-VL-72B~\cite{qwen25} model as the GEdit evaluation backbone.

For comparison, we construct prompts in the form ``make [object] \textit{slightly/mo\-derately/extremely} [attribute1]'' and ``\textit{slightly/moderately/extremely} [attribute2]'' to evaluate multi-attribute and continuous control in VAREdit and Qwen-Image. Concept-slider follows the same evaluation protocol as ATA.

\noindent\textbf{Implementation.}
All experiments were conducted in an identical environment for a fair comparison (Hardware: a single NVIDIA H100 GPU; Software: PyTorch 2.5.1, CUDA 12.2). We optimize Delta-Mod for 500 steps over 5 epochs with a batch size of 1 and use the original VAR objective to optimize Delta-Mod, \ie, the cross-entropy loss.

\begin{table}[t]
\caption{\textbf{Quantitative comparison of semantic alignment and content preservation}. We use $\Delta$VQA and Semantic Consistency (S.C.) to measure the semantic shift toward the target attribute. Image preservation is evaluated via the inverted LPIPS metric (I-LPIPS) and Perceptual Quality (P.Q.). Moreover, we leverage Disen. for the disentanglement performance in multi-attribute editing.}
\vspace{-20pt}
\begin{center}
\resizebox{\textwidth}{!}{%
\begin{tabular}{c|c|cc|ccc}
\toprule[1.5pt]
\multirow{2}{*}{Model} & \multirow{2}{*}{Intensity} & \multicolumn{2}{c|}{Control Dataset} & \multicolumn{3}{c}{GEdit} \\
\cmidrule(lr){3-4} \cmidrule(lr){5-7}
 & & $\Delta$VQA($\uparrow$) & I-LPIPS($\uparrow$) & S.C.($\uparrow$) & P.Q.($\uparrow$) & Disen.($\uparrow$) \\
 \midrule 
\multirow{3}{*}{Concept-slider} & Slight & 0.421 & 0.891 & 4.992 & 7.612 & 5.172 \\
 & Moderate & 0.513 & 0.847 & 4.826 & 7.261 & 5.018 \\
 & Extreme & 0.572 & 0.812 & 4.622 & 7.027 & 4.920 \\
 \midrule 
\multirow{3}{*}{Qwen-Image} & Slight & +3.82\% & \textbf{+4.12\%} & +7.13\% & +4.27\% & +5.02\% \\
 & Moderate & +5.31\% & \textbf{+5.44\%} & +6.20\% & +5.11\% & +3.86\% \\
 & Extreme & +7.01\% & +6.89\% & +6.34\% & \textbf{+6.18\%} & +3.42\% \\
 \midrule 
\multirow{3}{*}{VAREdit} & Slight & +5.71\% & +3.16\% & +6.68\% & +4.56\% & +4.88\% \\
 & Moderate & +6.30\% & +4.75\% & +5.72\% & \textbf{+5.75\%} & +4.52\% \\
 & Extreme & +6.95\% & +6.37\% & +5.51\% & +5.78\% & +4.35\% \\
  \midrule 
\multirow{3}{*}{\textbf{Ours}} & Slight & \textbf{+11.44\%} & +3.68\% & \textbf{+10.28\%} & \textbf{+4.82\%} & \textbf{+8.42\%} \\
 & Moderate & \textbf{+12.72\%} & +5.17\% & \textbf{+16.33\%} & +3.16\% & \textbf{+9.68\%} \\
 & Extreme & \textbf{+14.58\%} & \textbf{+7.29\%} & \textbf{+8.24\%} & +3.64\% & \textbf{+6.78\%} \\
\bottomrule[1.5pt]
\end{tabular}
}
\end{center}
\label{table_q}
\vspace{-15pt}
\end{table}

\subsection{Qualitative Evaluation}

Figure~\ref{fig:geneattr} presents side-by-side comparisons with Qwen-Image-Edit, VAREdit and Concept-slider on two challenging attribute transfer tasks: applying the \textit{apple-shaped} attribute to cakes and transferring the \textit{Sphynx} concept from sphynx cats to sheep.

\noindent\textbf{General attribute transfer.}
Our method generates smooth, continuous transformations while preserving object identity.
Baselines either perform object substitution (\eg, cake $\rightarrow$ apple) or fail to express the fine-grained attribute (\eg, hairlessness and wrinkles in the Sphynx concept).
These results demonstrate that our semantic directions capture \textbf{class-agnostic, high-level semantic properties} that existing methods cannot reliably transfer.

\begin{wraptable}{r}{0.48\textwidth}
\centering
\vspace{-35pt}
\caption{\textbf{User study.} Preference rates for Image Preservation (I.P.) and Prompt Adherence (P.A.).}
\label{tab:user_study}
\begin{tabular}{lcc}
\toprule
Model & I.P.($\uparrow$) & P.A.($\uparrow$) \\
\midrule
Concept-slider & 10\% & 6\% \\
Qwen-Image     & 30\% & 23\% \\
VAREdit        & 23\% & 27\% \\
\midrule
\textbf{Ours}  & \textbf{37\%} & \textbf{43\%} \\
\bottomrule
\end{tabular}

\end{wraptable}

\subsection{User Study}
To complement the above evaluation, we conduct a user study to assess human perception of the generated images. We follow the standard experimental paradigm from psychophysics, namely the two-alternative forced choice (2AFC) setting~\cite{motamed2025generative}. In our study, participants were given a questionnaire in which they viewed pairs of images showing continuous semantic control of a reference image.
For each trial, we show a reference image and two edited results produced by different methods under the same attribute prompt. 
Participants are asked to choose the result that better matches the target attribute while preserving the reference content.
The responses from 30 participants are summarized in Table~\ref{tab:user_study}. The results show a clear preference for the images generated by our method over those produced by the baselines.

\subsection{Quantitative Evaluation}
Table~\ref{table_q} reports quantitative results. 
Our method achieves the largest semantic shift towards the target attribute ($+11.44\%$ / $+12.72\%$ / $+14.58\%$ $\Delta$VQA) while maintaining competitive or superior image preservation (up to $+7.29\%$ I-LPIPS at extreme intensity) on controllable dataset.
Moreover, our approach outperforms existing models in semantic consistency and disentanglement performance in GEdit, particularly in complex multi-attribute editing scenarios (\eg, $+16.33\%$ Semantic Consistency and $+9.68\%$ Disentanglement).
Notably, our method requires neither LoRA optimization nor fine-tuning of VAR, yet outperforms instruction-based editors across all three control settings.

\subsection{Ablation Studies}

\begin{figure}[tb]
  \centering
  \includegraphics[width=\textwidth]{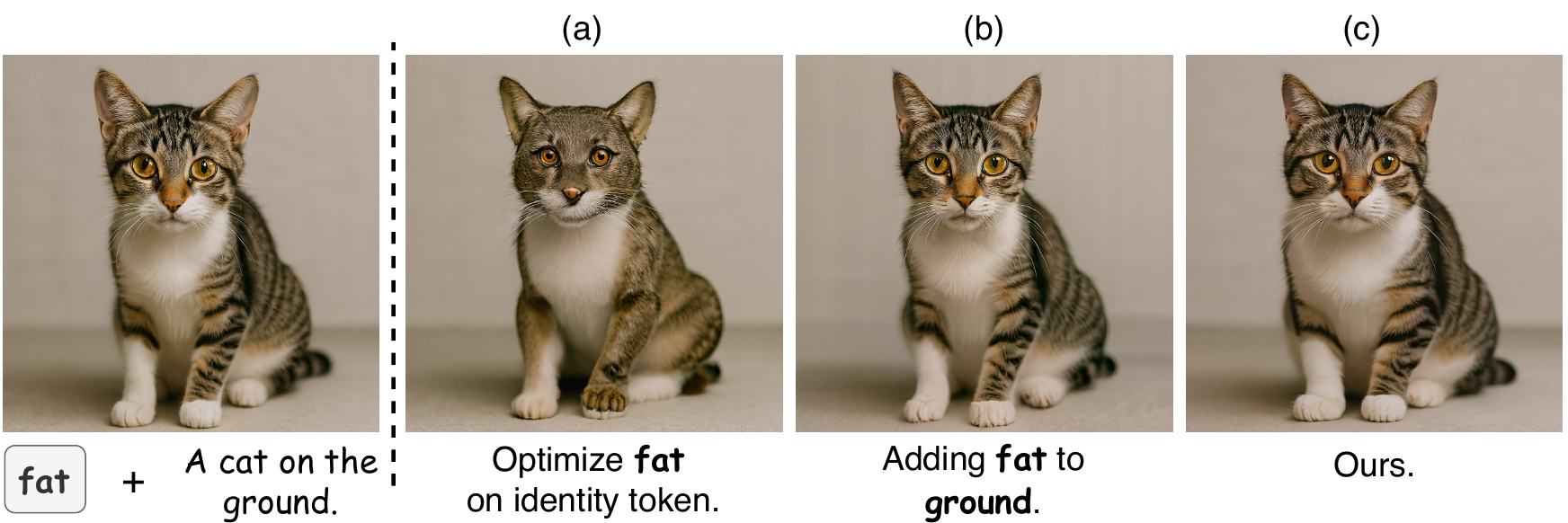}
  \caption{\textbf{Ablation study}. (a) Optimizing the attribute direction on the identity token `cat' instead of the attribute token `A' changes the subject identity in the generated image. (b) Adding the fat direction into a non-attribute token `ground' does not produce a fatter cat. (c) Our method can make a fatter cat while preserving identity.
  }
  \vspace{-10pt}
  \label{fig:ablation}
\end{figure}

We analyze the key components of our design:

\noindent\textbf{Injecting attribute directions to identity.}
Fig.~\ref{fig:ablation}(a) shows that adding attribute directions to identity tokens results in semantic conflicts and degraded edits.
This validates our identity–attribute separation strategy.

\noindent\textbf{Token-position dependence.}
In Fig.~\ref{fig:ablation}(b), applying a direction (\eg, \textit{fat}) to an unrelated token yields no meaningful change, showing that directions must be injected at their corresponding semantic positions.

\noindent\textbf{Full design.}
Fig.~\ref{fig:ablation}(c) illustrates that our full model produces clean, attribute-specific edits with no identity leakage.

\noindent\textbf{Training-free text-token baseline.}
To examine whether attribute directions can be obtained by direct text-token arithmetic, we compare ATA with a training-free baseline that uses the raw token difference in Eq.~\eqref{eq:sd}, \ie, $\mathrm{mean}(T_{\rm attr}^{\rm tar}) - T_{\rm attr}^{\rm src}$, as the semantic direction.
Both methods are evaluated under the same GEdit protocol, and we report mean scores in Table~\ref{tab:text_token_baseline}.
ATA consistently outperforms this baseline across all three metrics, suggesting that learned directions capture attribute factors more effectively than naive prompt-level token differences.

\begin{table}[tb]
  \centering
  \caption{\textbf{Comparison with a training-free text-token baseline.} We report mean scores under the same GEdit protocol. The baseline directly uses the token difference in Eq.~\eqref{eq:sd} as the semantic direction.}
  \label{tab:text_token_baseline}
  \resizebox{0.86\textwidth}{!}{
  \begin{tabular}{lccc}
    \toprule
    Method & Semantic Consistency ($\uparrow$) & Perceptual Quality ($\uparrow$) & Disentanglement ($\uparrow$) \\
    \midrule
    $\mathrm{mean}(T_{\rm attr}^{\rm tar}) - T_{\rm attr}^{\rm src}$ & 5.048 & 7.447 & 5.296 \\
    \textbf{ATA (Infinity-2B)} & \textbf{5.374} & \textbf{7.584} & \textbf{5.455} \\
    \bottomrule
  \end{tabular}
  }
  \vspace{-8pt}
\end{table}

\noindent\textbf{Comparison with Infinity-2B using Concept Slider.}
We further compare with an Infinity-2B baseline trained using the Concept Slider strategy.
Specifically, we perform LoRA tuning for attribute control on the Control Dataset while keeping the same Infinity-2B backbone as ATA.
As shown in Table~\ref{tab:control_dataset_ablations}, ATA with original prompts achieves higher mean $\Delta$VQA and mean I-LPIPS than Infinity-2B with Concept Slider.
This result suggests that the advantage of ATA is not solely due to the backbone and that its token-direction formulation provides more effective attribute control.

\noindent\textbf{Prompt sensitivity.}
ATA is not designed around a fixed prompt template.
To evaluate its sensitivity to prompt wording, we perturb prompts on the Control Dataset using natural variations, including template changes (\eg, ``\ttb{a cat}'' $\rightarrow$ ``\ttb{a photo of a cat}''), lexical paraphrases (\eg, ``\ttb{a cat}'' $\rightarrow$ ``\ttb{a kitty}''), and context deletion (\eg, ``\ttb{a cat on the floor}'' $\rightarrow$ ``\ttb{a cat}'').
As shown in Table~\ref{tab:control_dataset_ablations}, these perturbations lead to modest drops from the original-prompt ATA setting, suggesting that ATA remains effective under common prompt variations while leaving further robustness improvements for future work.

\begin{table}[tb]
  \centering
  \caption{\textbf{Ablations on the Control Dataset.} ATA with original prompts serves as the shared reference for comparison with Infinity-2B using Concept Slider and for the prompt sensitivity study.}
  \label{tab:control_dataset_ablations}
  \resizebox{0.86\textwidth}{!}{
  \begin{tabular}{llcc}
    \toprule
    Analysis & Method / Setting & Mean $\Delta$VQA ($\uparrow$) & Mean I-LPIPS ($\uparrow$) \\
    \midrule
    \multirow{2}{*}{Same-backbone comparison}
    & Infinity-2B with Concept Slider & 0.532 & 0.879 \\
    & ATA (Infinity-2B, original prompts) & \textbf{0.567} & \textbf{0.895} \\
    \midrule
    Prompt sensitivity
    & ATA (Infinity-2B, perturbed prompts) & 0.541 & 0.864 \\
    \bottomrule
  \end{tabular}
  }
  \vspace{-8pt}
\end{table}

\begin{figure}[tb]
  \centering
  \includegraphics[width=\textwidth]{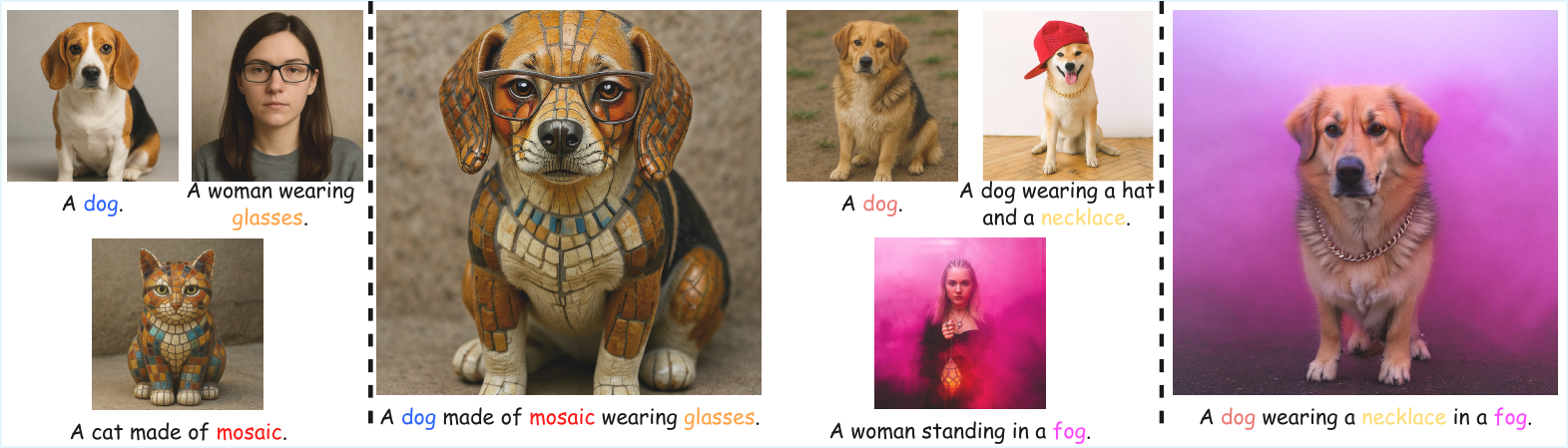}
  \caption{\textbf{Identity–attribute compositionality}. We extract identity directions from three reference images and recombine them with various attribute directions. Our method preserves the identity of the chosen reference while expressing attributes such as \textit{glasses}, \textit{hat}, \textit{necklace}, or \textit{fog}, illustrating clean semantic disentanglement.
  }
  \label{fig:compos}
  \vspace{-5pt}
\end{figure}

\begin{figure}[tb]
  \centering
  \includegraphics[width=\textwidth]{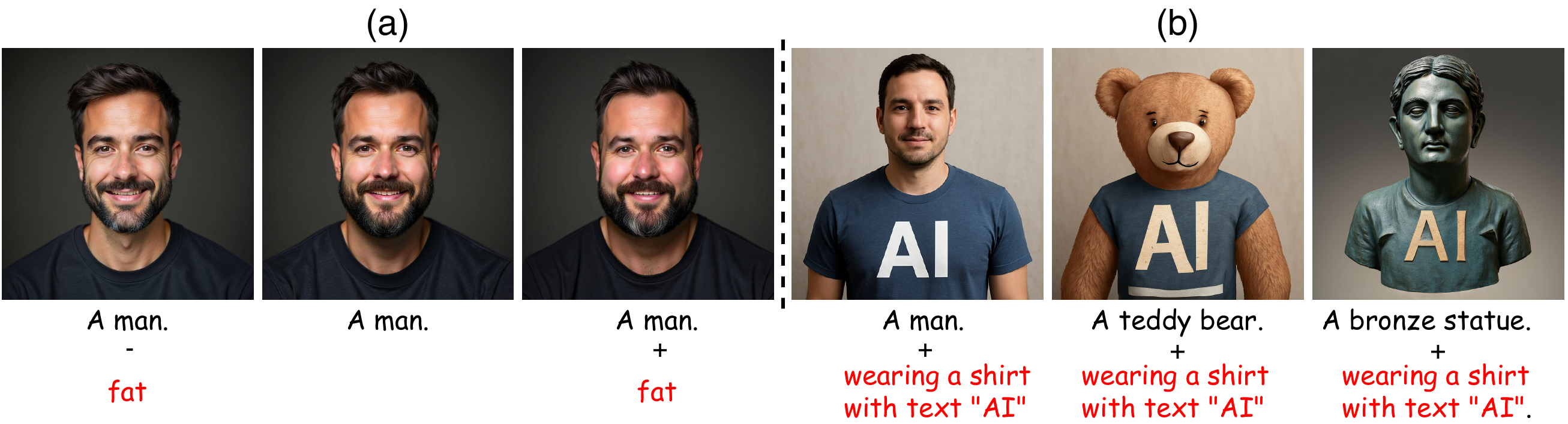}
  \caption{\textbf{Advanced Editing Capabilities}. (a) Our model can both intensify (+ fat) and diminish (– fat) the same attribute, while preserving the subject’s identity and overall visual coherence. (b) Beyond simple word-level attributes, ATA enables the transfer of complex, composite attributes across distinct object categories. For example, the attribute ``\ttb{wearing a shirt with the text AI}'' can be consistently applied to a human, a teddy bear, and a bronze statue.
  }
  \vspace{-10pt}
  \label{fig:advanced}
\end{figure}

\subsection{Advanced Editing Capabilities}

We further demonstrate several advanced semantic control scenarios.

\noindent\textbf{Identity–attribute compositionality.}
Fig.~\ref{fig:compos} demonstrates that our method learns image-specific identity directions, which can be recombined with arbitrary attribute directions at inference time. 
Given three reference identities (a dog, a woman, and a mosaic cat), our model extracts separate identity directions and applies them to new scenes while controlling attributes such as \textit{glasses}, \textit{hat}, \textit{necklace}, or \textit{fog}. 
Across all cases, the synthesized images preserve the target identity while faithfully expressing the desired attributes, confirming that identity and attribute semantics are disentangled and compositional in our representation.

\noindent\textbf{Bidirectional and continuous control.}
Fig.~\ref{fig:advanced}~(a) demonstrates that our semantic directions support positive and negative traversal (\eg, $-$fat $\rightarrow$ original $\rightarrow$ $+$fat),
along the semantic attribute axis.

\noindent\textbf{Control of complex attributes.}
For simple attributes (\eg, fatness, emotion), a single word can describe a specific aspect of an object. 
However, real-world attributes are often more complex and cannot be captured by one or two words. 
Interestingly, our method can also transfer such complex semantic concepts, which may involve multiple attributes or even full descriptive phrases. 
As shown in Fig.~\ref{fig:advanced}~(b), beyond simple attributes, we can leverage complex descriptions as abstract attributes for image editing, further improving our flexibility.

\begin{figure}[tb]
  \centering
  \includegraphics[width=\textwidth]{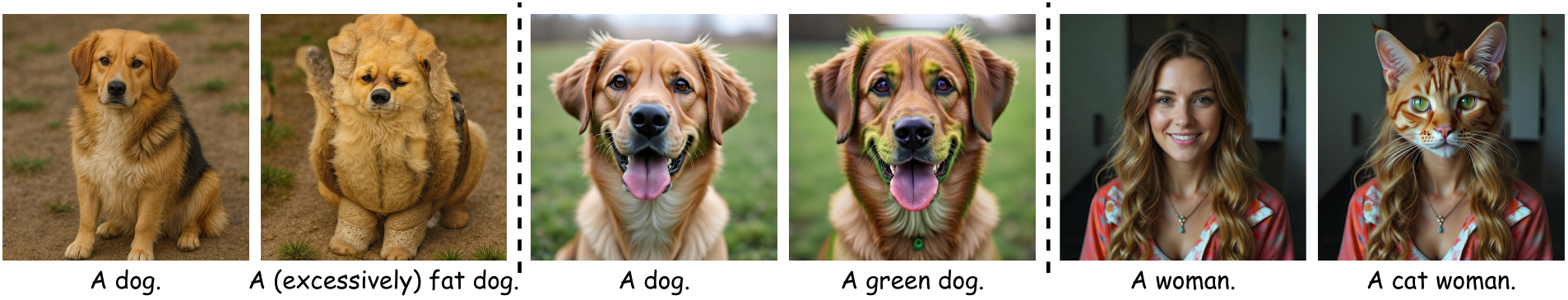}
  \caption{\textbf{Failure cases}. (a) There is a bound on attribute strength; beyond this bound, artifacts appear in the generated images; (b) The method is less effective for certain particular attributes (\eg, color); (c) ATA is limited by the pre-trained generative prior of the backbone. 
  }
  \vspace{-15pt}
  \label{fig:limitation}
\end{figure}

\subsection{Limitation and Discussion}
Although our model can transfer general attributes and support continuous control, the editing capacity is ultimately constrained by the generative prior of VAR.
Very strong attribute strengths may introduce artifacts or affect unrelated regions (see Fig.~\ref{fig:limitation}~(a)).
Additionally, global color transformations remain challenging (see Fig.~\ref{fig:limitation}~(b)). We hypothesize that this behavior is related to VAR’s hierarchical generation mechanism. VAR predicts tokens from coarse to fine, and its localized attributes may be spatially or hierarchically entangled across neighboring tokens/scales. Such entanglement makes it difficult to disentangle these attributes under our ATA design.
Moreover, due to the limited world knowledge of the base model, ATA cannot transfer concepts that the base model itself does not understand. For example, attempting to transfer the concept of a ``cat'' onto a woman does not result in a woman with cat-like styling, as this type of cross-domain semantic mapping is not well supported by the backbone, as shown in Fig.~\ref{fig:limitation}~(c). Future work will explore more autoregressive frameworks for a comprehensive validation.

\section{Conclusion}
Autoregressive text-to-image models have recently demonstrated strong performance as a unified framework for image generation. However, achieving fine-grained semantic control remains challenging due to the attribute entanglement and the misalignment between textual and visual representations. In this work, we have presented Attribute Token Arithmetic (ATA), a token-level semantic direction manipulation method built upon the VAR backbone. ATA enables disentangled and continuous attribute control by performing arithmetic operations directly in the VAR token space after a few optimization epochs. Experiments demonstrate the effectiveness and generalizability of ATA, showing consistent improvements over existing approaches for controllable generation.

\section*{Acknowledgements}
This research was supported in part by the Australian Government through the Australian Research Council (ARC) Discovery Project DP260100218.


\bibliographystyle{splncs04}
\bibliography{main}
\end{document}